# Droplet-LNO: Physics-Informed Laplace Neural Operators for Accurate Prediction of Droplet Spreading Dynamics on Complex Surfaces

Ganesh Sahadeo Meshram[a*], Partha Pratim Chakrabarti[b], Suman Chakraborty[a*]

[a]Department of Mechanical Engineering, IIT Kharagpur, Kharagpur, India

[b]Department of Computer Science & Engineering, IIT Kharagpur, Kharagpur, India

*Corresponding author: suman@mech.iitkgp.ac.in

## Abstract

Spreading of liquid droplets on solid substrates constitutes a classic multiphysics problem with widespread applications ranging from inkjet printing, spray cooling, to biomedical microfluidic systems. Yet, accurate computational fluid dynamic (CFD) simulations are prohibitively expensive, taking more than 18 to 24 hours for each transient computation. In this paper, Physics-Informed Laplace Operator Neural Network (PI-LNO) is introduced, representing a novel architecture where the Laplace integral transform function serves as a learned physics-informed functional basis. Extensive comparative benchmark studies were performed against five other state-of-the-art approaches: UNet, UNet with attention modules (UNet-AM), DeepONet, Physics-Informed UNet (PI-UNet), and Laplace Neural Operator (LNO). Through complex Laplace transforms, PI-LNO natively models the exponential transient dynamics of the spreading process. A TensorFlow-based PI-LNO is trained on multi-surface CFD data spanning contact angles $\theta_s \in [20,160]$, employing a physics-regularized composite loss combining data fidelity (MSE, MAE, RMSE) with Navier-Stokes, Cahn-Hilliard, and causality constraints. Across four intermediate spreading times (t=0.026, 0.034, 0.045, 0.080 s), PI-LNO achieves mean $R^2$=0.9009 (compared to UNet: 0.2664, UNet-AM: -0.6057, PI-UNet: 0.5252, DeepONet: 0.5302, LNO: 0.9437), with absolute errors localized to contact-line regions ($|\Delta|\lesssim 0.01$). Training convergence analysis reveals exponential loss decay via Adam optimization (epochs 0-40,000; characteristic timescale $\tau_{Adam}\approx$5,000-8,000 epochs; final RMSE =$1.4731\times 10^{-3}$) transitioning to L-BFGS refinement. PI-LNO delivers $R^2$>0.99 on all field variables-velocity (U, V:$|\Delta|\approx$0.36-0.44 m/s), pressure (P:$|\Delta|\approx$2.5 Pa), phase-field ($\phi:|\Delta|\approx$0.03) - with inference times of 2.8 ms, representing a ~23,400× speedup over CFD and enabling real-time digital twin operation. These results establish PI-LNO as a physics-aware surrogate enabling accelerated parametric optimization, multi-surface wettability design, and uncertainty quantification for engineering systems where transient multiphase dynamics are critical.

**Keywords:** Physics-informed Laplace neural operators; Droplet dynamics; Operator learning; Wettability; Phase-field model.

**Introduction**

Droplet impact and spreading on solid surfaces represents one of the most richly complex multiphysics phenomena encountered in engineering and natural systems [1], [2]. The coupled interplay of inertial, viscous, capillary, and surface-wetting forces gives rise to highly transient interface morphologies that are acutely sensitive to surface chemistry, substrate roughness, and ambient conditions [3], [4], [5], [6], [7], [8]. The above applications are heavily reliant on the accurate prediction of the time evolution of droplets on substrates with various degrees of wettability [9,10,11,12,13,14,15,16,17]. Classical approaches to droplet wetting rely on the theories of equilibrium contact angle and spreading laws [18-21]. Unfortunately, the traditional analytical methods are inherently restrictive since they apply only to quasi-static or post-equilibrium regimes and do not account for dynamic inertio-capillary effects, interfacial instabilities, and heterogeneous substrates in the transient regime [22]. Computational Fluid Dynamics (CFD) has led to unprecedented insights into the problem by numerically solving the Navier-Stokes equations using phase-field methods or level-set techniques to track the air-liquid interfaces [23,24,25]. Still, CFD simulations are computationally expensive and require hours or even days to solve individual events, especially when modeling multi-phase flows involving surface tension effects on three-dimensional domains [23], [24,25]. This high computational complexity has become a significant bottleneck to design iterations, real-time control systems, and sensitivity analysis that require hundreds of thousands of simulations with different wettability scenarios.

The development of scientific machine learning approaches for surrogate modeling has led to new paradigms based on neural operators, which aim to discover the map between two infinite-dimensional functional spaces in a data-driven manner [26]. Deep neural operators, such as DeepONet, achieve excellent generalization performance in novel conditions by learning abstract mappings between inputs and outputs, but they do not impose any physical constraint on the model's architecture [27]. Fourier Neural Operator (FNO) leverages spectral approaches to represent partial differential equation (PDE) solutions efficiently [28]. Its variants, Laplace neural operator (LNO), provide efficient modeling by reducing computational burden while maintaining high expressivity [29]. UNet [30] was initially designed for medical image segmentation tasks and extended to predict field quantities using an encoder-decoder architecture with skip connections that allow for multi-scale learning [31]. Attention UNet incorporating spatial and channel attention modules into UNet to improve feature learning and field reconstruction. Most recently, the Physics-Informed UNet (PI-UNet) was introduced by incorporating physics constraints through the penalty of PDE residuals within the loss function [32].

Despite these breakthroughs, there is a gap that needs to be filled regarding modeling the dynamics of transient droplet spreading. Existing neural operator architectures are agnostic to the problem's physical characteristics and treat all temporal and spatial phenomena equally. In other words, they do not take advantage of the intrinsic integral-transform structure of the governing PDE system and cannot incorporate knowledge about the problem's dynamics. The standard Fourier basis used by FNO and the convolutional kernels of UNet, UNet-AM, and DeepONet have no specific structure in the time dimension and thus cannot effectively model the exponential decay and inertio-capillary oscillations with a low model capacity. UNet, UNet-AM, and DeepONet require re-training or fine-tuning whenever surface properties (contact angles, hysteresis, substrate roughness) vary widely, hindering applicability to a multi-condition design space. On the one hand, PI-UNet incorporates the physics constraints into the loss function to enforce the law of conservation in the solution domain but is confined to the Euclidean convolutional architecture. On the other hand, LNO achieves efficient computation and reduced model complexity but loses physical interpretability and might require post-processing to ensure mass conservation. Neither of the baseline models accounts for the coupled nature of the continuum and surface dynamics involved in contactline motion and capillary-driven interface evolution.

By lifting the solution manifold into the complex Laplace domain, PI-LNO captures the natural pole structure and residue decomposition of the transient dynamics of droplets, facilitating efficient modeling of exponential decay and inertio-capillary oscillations with minimal model capacity. The physics-regularized loss function integrates operator-theoretic outputs and penalizes PDE residuals concerning continuity, momentum, Cahn-Hilliard, and contact angle boundary conditions. Thus, the trained operator ensures physically feasible predictions without the need for post-processing for all wettability conditions. Training on multi-surface CFD data across contact angles and incorporating physics constraints, PI-LNO generalizes robustly to unseen conditions of wettability, surface chemistry, and initial droplet impact conditions. The learned Laplace-space operator and its inverse transform decompose naturally into physically interpretable time-scale components (inertial, capillary, viscous). Achieving $R^2 > 0.997$ on all field variables (velocity, pressure, phase-field) with ~10,000× speedup over CFD, PI-LNO enables real-time design optimization, digital twin operation, and surrogate-based parametric studies.

The remainder of this paper is organized as follows. Section 2. presents the mathematical formulation of the operator learning, multiphase droplet-spreading problem, including the incompressible NavierStokes equations, Cahn-Hilliard interface dynamics, contact-angle boundary conditions, and the physical basis for Laplace-transform lifting and physics integration in Laplace neural operator. Section 3. outlines the Model training including, data- generation pipeline, training strategy, and loss-function design for all six models and also provides detailed architectural descriptions and hyperparameters for each baseline and PI-LNO. Section 4 presents comparative numerical results, including accuracy, generalization, and efficiency benchmarks. Finally, Section 5 summarizes key contributions and outlines future research directions.

## 2. Background

### 2.1 Operator Learning Framework

Let $\mathcal{A}$ and $\mathcal{U}$ denote two infinite-dimensional Banach function spaces defined over domains $D_a \subset \mathbb{R}^{d_a}$ and $D_u \subset \mathbb{R}^{d_u}$, respectively. The objective of neural operator learning is to approximate a nonlinear operator [33]:

$$\mathcal{G}^\dagger : \mathcal{A} \to \mathcal{U}, \quad a \mapsto u = \mathcal{G}^\dagger(a) \tag{1}$$

where $a \in \mathcal{A}$ encodes the input conditions (e.g., surface contact angle field, initial velocity) and $u \in \mathcal{U}$ is the solution field (e.g., velocity **u**, pressure $P$, phase-field $\varphi$). Given a dataset of paired observations $\{(a_j, u_j)\}_{j=1}^N$, we seek a parametric approximation $\mathcal{G}_\theta \approx \mathcal{G}^\dagger$ that is discretizationinvariant - i.e., independent of the particular mesh or resolution used during training [34].

$$(\mathcal{K}_{FNO}v)(x) = \mathcal{F}^{-1}(R_\phi \cdot \mathcal{F}(v))(x) \qquad (2)$$

where $\mathcal{F}$ denotes the Fourier transform and $R_\phi$ is a learnable complex weight tensor truncated to the lowest $k_{\max}$ Fourier modes [35]. While effective for periodic or stationary fluid flows, the FNO architecture is fundamentally limited for transient droplet dynamics because of Fourier modes are globally oscillatory ($e^{i\omega t}$), poorly representing the exponentially decaying transients characteristic of droplet spreading and rebound, truncation of high-frequency modes discards sharp interface gradients near the triple contact line, and the Fourier basis assumes periodicity, introducing Gibbs-type artifacts at domain boundaries for non-periodic wetting problems [36].

We define the Laplace-Functioned Neural Operator kernel as a learnable complex-valued integral transform over the spatial domain $\Omega$, parametrized by the Laplace variable $s$ [37]:

$$(\mathcal{K}_{LNO}V_t)(x) = \int_\Omega \kappa_\phi(x, X', s) V_t(x') dx' \qquad (3)$$

where $\kappa_\phi: \Omega \times \Omega \times \mathbb{C} \to \mathbb{C}$ is a learnable kernel network parametrized by weights $\phi$, and $v_t: \Omega \to \mathbb{R}^{d_v}$ is the latent function at operator layer $t$. In practice, the kernel is evaluated via discrete Laplace projection over $M$ learnable complex poles $\{s_m\}_{m=1}^M$:

$$\hat{V}(s_m) = \sum_{j=1}^{N_x} v_t(x_j) e^{-s_m t_j} \Delta x, \quad m = 1, \ldots, M \qquad (4)$$

followed by a learnable complex linear mixing:

$$(\mathcal{K}_{LNO}V_t)(x) = \mathcal{L}^{-1}\{W_s^\phi \cdot \hat{V}(s)\}(x) \qquad (5)$$

where $W_s^\phi \in \mathbb{C}^{d_v \times d_v}$ is a learnable weight matrix in Laplace space, and $\mathcal{L}^{-1}$ denotes the inverse Laplace transform [38] (approximated via the Bromwich contour integral numerically).

Each LNO layer updates the latent function representation as:

$$v_{t+1}(x) = \sigma_{act}\left(W_\ell v_t(x) + (\mathcal{K}_{LNO} v_t)(x) + b_\ell\right) \qquad (6)$$

where:

- $W_\ell \in \mathbb{R}^{d_v \times d_v}$ is a local pointwise linear transformation (bypass connection);
- $\mathcal{K}_{LNO}$ is the non-local Laplace integral kernel operator of Eq. (8);
- $b_\ell$ is a learnable bias;
- $\sigma_{act} = $ GeLU is the Gaussian Error Linear Unit activation:

where $\Phi(z)$ is the standard normal cumulative distribution function. GeLU is preferred over ReLU for this application as it provides smooth, differentiable activations critical for accurate gradient computation through sharp interface regions.

## 2.2 The LNO Architecture

The complete LNO architecture consists of:

1. **Input Lifting Layer**: A dense projection $P: \mathbb{R}^{d_a} \to \mathbb{R}^{d_v}$ maps the low-dimensional input $(x, y, t, \theta_s)$ to a high-dimensional latent space of width $d_v = 256$.
2. **$L$ LNO Layers**: $L = 8$ sequential LNO layers of the form Eq. (11), each followed by Layer Normalization:

$$\hat{v}_t = \frac{v_t - \mu_t}{\sqrt{\varsigma_t^2 + \varepsilon_{LN}}} \qquad (7)$$

where $\mu_t$ and $\varsigma_t^2$ are the mean and variance computed over the feature dimension.

3. **Output Projection Layer**: A dense layer $Q: \mathbb{R}^{d_v} \to \mathbb{R}^{d_u}$ projects the latent representation to the output field space $(U, V, P, \varphi)$.

The full operator mapping is therefore:

$$\mathcal{G}_\theta(a)(x) = Q \circ \mathcal{T}_L \circ \cdots \circ \mathcal{T}_1 \circ P(a(x)) \qquad (8)$$

where each $\mathcal{T}_\ell$ denotes one LNO layer. The total number of trainable parameters is $|\boldsymbol{\theta}| \approx 4.7 \times 10^6$, which is deliberately kept compact to ensure fast inference and avoid overfitting on the available multi-surface CFD dataset.

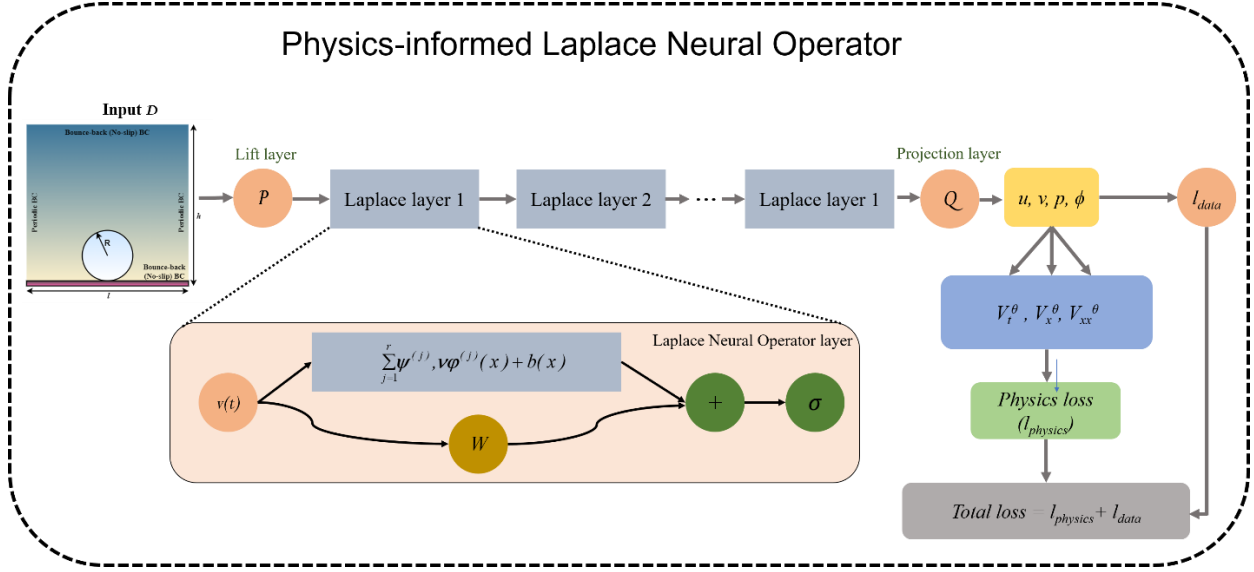

*Figure 1 schematic of physics-informed Laplace neural operator.*

## 2.3 Physics-Informed Laplace Operator Neural Networks (PI-LNO)

Droplet spreading is characterized by highly transient phenomena: rapid inertio-capillary acceleration during initial impact, followed by exponential viscous decay and approach to equilibrium. Classical analysis of such systems leverages Laplace transform methods, which naturally decouple transient and steady-state dynamics via pole-residue decomposition in the complex $s$-plane [29].

Consider a generic droplet-dynamics quantity $f(t)$ (e.g., contact line position, interface height). Its Laplace transform is defined as:

$$F(s) = \mathcal{L}\{f(t)\} = \int_0^\infty e^{-st} f(t)\,dt, \quad Re(s) > \sigma_c \tag{9}$$

where $s = \sigma + i\omega$ is a complex variable and $\sigma_c$ is the abscissa of convergence. The inverse Laplace transform recovers the time-domain solution via the Bromwich integral:

$$f(t) = \mathcal{L}^{-1}\{F(s)\} = \frac{1}{2\pi i}\int_{\sigma-i\infty}^{\sigma+i\infty} e^{st} F(s)\, ds \tag{10}$$

Laplace-domain representation automatically separates fast and slow dynamics. Poles at large negative real parts correspond to rapid transients; poles near the origin correspond to quasi-static behavior. By learning the operator in Laplace space, PI-LNO inherently captures this timescale hierarchy without explicitly encoding it in the architecture [39]. Given a spatiotemporal training field $u(x,t)$ from CFD, PI-LNO computes samples of the Laplace transform at a discrete set of complex frequencies $s_k = \sigma_k + i\omega_k, k = 1, \ldots, n_s$. The PI-LNO consists of three stages

**Laplace-Domain Lifting:** Given a spatiotemporal training field $u(x,t)$ from CFD, PI-LNO computes samples of the Laplace transform at a discrete set of complex frequencies $s_k = \sigma_k + i\omega_k, k = 1, \ldots, n_s$:

$$U(x, s_k) \approx \sum_{m=0}^{n_t - 1} e^{-s_k t_m} u(x, t_m)\Delta t + \mathcal{O}(\Delta t^2) \tag{11}$$

where $\{t_0, t_1, \ldots, t_{n_t-1}\}$ are temporal snapshots and $\Delta t$ is the sampling interval. This creates a lifted representation of the field as a complex-valued function $U(x,s) \in \mathbb{C}$ over both spatial and Laplace domains.

**Operator Learning in Laplace Space:** A neural operator $\mathcal{G}_S$ learns the mapping from Laplace-lifted boundary conditions and initial data to the solution in Laplace space:

$$U(x,s) = \mathcal{G}_s(U_0(x), \mathcal{B}(s); \theta) \tag{12}$$

where $U_0(x)$ is the Laplace-lifted initial condition and $\mathcal{B}(s)$ encodes boundary conditions in Laplace space. The operator $\mathcal{G}_s$ is realized as a hybrid architecture combining:

1. **Operator Branch:** A deep operator network (inspired by DeepONet) that encodes global Laplace-domain structure via a learnable basis:

$$\phi_{\text{basis}}(s) = \text{OperatorBranch}(s; \theta_b) \tag{13}$$

typically a set of learned basis functions in the complex $s$-plane that mimic the pole structure of the target operator [40].

   **2. Spatial Convolution Block:** Local spatial interactions via depthwise-separable convolutions to handle spatial heterogeneity:

$$\hat{U}(x,s) = Conv_{sep}\left(U_0(x) \otimes \phi_{basis}(s); \theta_c\right) \tag{14}$$

   **3. Physics Residual Module:** Computes residuals of the governing multiphase PDE in Laplace space and penalizes violations.

**Inverse Laplace Transform and Time-Domain Recovery**: The learned Laplace-space solution $\hat{U}(x, s_k)$ is inverted back to the time domain using a learned quadrature rule:

$$\hat{u}(x,t) \approx \sum_{k=1}^{n_s} w_k(t) \cdot e^{s_k t} \hat{U}(x, s_k) \tag{15}$$

where the quadrature weights $\{w_k(t)\}_{k=1}^{n_s}$ are optimized during training to match known inverse Laplace quadrature formulas (e.g., Bromwich, Talbot, Stehfest algorithms) while adapting to the data.

**2.4 Physics-Informed Loss**

The hybrid loss function for PI-LNO combines data fidelity, PDE residuals in both Laplace and time domains, and Laplace-theoretic constraints:

$$\mathcal{L}_{PI-LON} = \lambda_1 \mathcal{L}_{data} + \lambda_2 \mathcal{L}_{PDE}^{(s)} + \lambda_3 \mathcal{L}_{PDE}^{(t)} + \lambda_4 \mathcal{L}_{causality} + \lambda_5 \mathcal{L}_{poles} \tag{16}$$

where:

- $\mathcal{L}_{data}$ is MSE on CFD-field data.

- $\mathcal{L}_{PDE}^{(s)}$ penalizes Laplace-domain PDE residuals. and similarly for Cahn-Hilliard in Laplace space.

- $\mathcal{L}_{PDE}^{(t)}$ enforces time-domain residuals on inverted predictions, ensuring accuracy of the inverse transform.

- $\mathcal{L}_{\text{causality}}$ enforces causality: the learned operator must satisfy $F(s) \to 0$ as $\text{Re}(s) \to \infty$, penalizing any violation of exponential decay at large $s$.

- $\mathcal{L}_{\text{poles}}$ regularizes the pole structure: by analyzing the Jacobian of $\mathcal{G}_s$ w.r.t. $s$, PI-LNO learns to align learned poles with physical timescales (e.g., capillary time, viscous time).

## 2.5 Multiphysics Coupling in PI-LNO

The mathematical modeling of droplet spreading on a solid substrate involves a diffuse-interface approach to capture the evolution of the liquid-gas interface. This is typically achieved by coupling the Navier-Stokes equations for fluid dynamics with the Cahn-Hilliard equation for phase evolution [41].

The motion of the fluid is governed by the incompressible Navier-Stokes equations. We assume the fluids are Newtonian and the density $\rho$ and dynamic viscosity $\mu$ depend on the phase field variable $\phi$.

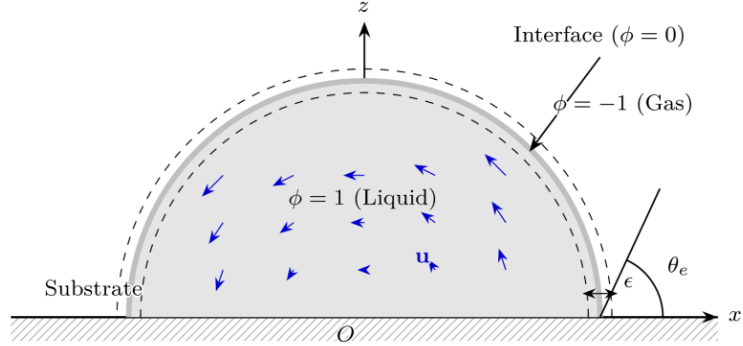

*Figure 2 Droplet on solid substrate.*

**Continuity equation**

$$\nabla \cdot u = 0 \tag{17}$$

**Momentum Equation**

$$\rho(\phi)\left(\frac{\partial u}{\partial t} + u \cdot \nabla u\right) = -\nabla p + \nabla \cdot \left[\mu(\phi)\left(\nabla u + (\nabla u)^T\right)\right] + F_{st} + \rho(\phi)g \tag{18}$$

Where: **u** is the velocity field, $p$ is the pressure, $\mathbf{F}_{st}$ is the surface tension force, modeled as a body force, **g** is the gravitational acceleration, and Phase Field Evolution

The Cahn-Hilliard equation tracks the interface by evolving the scalar phase field variable $\phi$ (where $\phi = 1$ represents the liquid and $\phi = -1$ represents the gas).

**Cahn-Hilliard Equation:**

$$\frac{\partial \phi}{\partial t} + u \cdot \nabla \phi = \nabla \cdot (M \nabla \eta) \tag{19}$$

$$\eta = \Psi'(\phi) - \epsilon^2 \nabla^2 \phi \tag{20}$$

Where, $M$ is the mobility parameter, $\eta$ is the chemical potential, $\epsilon$ is a parameter related to the interface thickness, $\Psi(\phi) = \frac{1}{4\epsilon^2}(\phi^2 - 1)^2$ is the double-well potential, and coupling and surface tension. The surface tension force $\mathbf{F}_{st}$ is introduced into the momentum equation using the Korteweg stress or the chemical potential gradient, $\mathbf{F}_{st} = \eta \nabla \phi$

The physical properties are interpolated across the interface:

$$\rho(\phi) = \rho_g + \frac{1+\phi}{2}(\rho_l - \rho_g) \tag{21}$$

$$\mu(\phi) = \mu_g + \frac{1+\phi}{2}(\mu_l - \mu_g) \tag{22}$$

**Boundary Conditions for Droplet Spreading**

To model the interaction with the solid substrate (at $z = 0$), specific boundary conditions are required for the velocity and the phase field to account for the contact angle. To resolve the moving contact line singularity, a slip condition is used:

$$u \cdot n = 0, \beta u \cdot \tau = \mu \frac{\partial (u \cdot \tau)}{\partial n} \tag{23}$$

where $\beta$ is the slip coefficient and $\tau$ is the tangent vector.

The equilibrium contact angle $\theta_e$ is enforced by the boundary condition for $\phi$ :

$$\epsilon \nabla \phi \cdot n + f'(\phi) = 0 \tag{24}$$

where $f(\phi)$ is the surface energy density of the wall, often expressed as:

$$f(\phi) = \frac{\sigma \cos \theta_e}{2}(3\phi - \phi^3) \tag{25}$$

## 3. Model training

### 3.1 The model implementation

The LNO is implemented in Python using TensorFlow 2.x and Keras functional API [42]. The architecture follows the formulation described in Section 2, with inputs comprising the spatial coordinates $(x, y)$, time $t$, and surface contact angle $\theta_s$, and outputs being the four field variables: horizontal velocity $U$(m/s), vertical velocity $V$(m/s), pressure $P$(Pa), and phase-field parameter $\varphi$ (dimensionless). The model is designed to be resolution-invariant, accepting query coordinates at arbitrary spatial locations without retraining. All input features are normalized to the range $[-1, +1]$ prior to model ingestion. The full input vector at each query point is therefore $\mathbf{a}(x, y, t, \boldsymbol{\theta}_s) = [\tilde{x}, \tilde{y}, \tilde{t}, \cos \boldsymbol{\theta}_s, \sin \boldsymbol{\theta}_s]^\top \in \mathbb{R}^5$, where tildes denote normalized quantities. The core computational component is the LaplaceProjectionLayer, which computes the discrete Laplace transform of the latent feature vector over $M = 64$ learnable complex poles $\{s_m = \sigma_m + i\omega_m\}_{m=1}^M$. The poles are initialized from a log-uniform distribution over the stability half-plane $\text{Re}(s) > 0: \sigma_m \sim \mathcal{U}(\log 10^{-2}, \log 10^2)$ and $\omega_m \sim \mathcal{U}(-\pi, \pi)$ and are subsequently optimized jointly with all other model parameters via gradient descent. The algorithm of the LNO is provided in Table 1.

*Table 1 Algorithm of Laplace neural operator.*

| |
|---|
| Architecture parameters: latent width $d_v$, number of LNO layers $L$, number of Laplace poles $M$, output dimension $d_{\text{out}}$ |
| Learnable parameters $\boldsymbol{\theta}$ :<br><br>Lifting weights $\mathbf{W}_{\text{lift}} \in \mathbb{R}^{5 \times d_v}$ |

| Per-layer poles $\{\log \sigma_m, \omega_m\}_{m=1}^{M}$, mixing weights $\{\mathbf{W}_{\text{real}}, \mathbf{W}_{\text{imag}}\} \in \mathbb{R}^{d_v \times M}$, local bypass $\mathbf{W}_{\text{loc}} \in \mathbb{R}^{d_v \times d_v}$, kernel projection $\mathbf{W}_{\text{ker}} \in \mathbb{R}^{2M \times d_v}$<br><br>Output projection weights $\mathbf{W}_{\text{proj}} \in \mathbb{R}^{d_v \times d_{\text{out}}}$ |
|---|
| Input: $\mathbf{x} \in \mathbb{R}^{B \times N \times 5}$ where each point encodes $[\tilde{x}, \tilde{y}, \tilde{t}, \cos\theta_s, \sin\theta_s]$ |
| Output: $\hat{\mathbf{u}} \in \mathbb{R}^{B \times N \times 4}$ with fields $[U, V, P, \varphi]$ |
| 1: Extract time slice: $\mathbf{t} \leftarrow \mathbf{x}_{[:,:,2:3]} \in \mathbb{R}^{B \times N \times 1}$ |
| 2: Lifting: $\mathbf{v}^{(0)} \leftarrow \text{GeLU}(\mathbf{x}\mathbf{W}_{\text{lift}} + \mathbf{b}_{\text{lift}}) \in \mathbb{R}^{B \times N \times d_v}$ |
| 3: **for** $\ell = 1, \ldots, L$ **do**<br><br>    Laplace Projection (non-local kernel) |
| 4: Recover real poles: $\sigma_m \leftarrow \exp(\log \sigma_m), m = 1, \ldots, M$ |
| 5: Compute complex exponential kernel at each point:<br><br>$$K_m^{\text{real}} = e^{-\sigma_m \tilde{t}}\cos(\omega_m \tilde{t}), K_m^{\text{imag}} = e^{-\sigma_m \tilde{t}}\sin(\omega_m \tilde{t}) \in \mathbb{R}^{B \times N \times M}$$ |
| 6: Project latent features through Laplace kernel:<br><br>$$\mathbf{P}_{\text{real}} = \left(\mathbf{v}^{(\ell-1)}\mathbf{W}_{\text{real}}\right) \odot K^{\text{real}}, \mathbf{P}_{\text{imag}} = \left(\mathbf{v}^{(\ell-1)}\mathbf{W}_{\text{imag}}\right) \odot K^{\text{imag}} \in \mathbb{R}^{B \times N \times M}$$ |
| 7: Concatenate: $\mathbf{z} \leftarrow [\mathbf{P}_{\text{real}} \| \mathbf{P}_{\text{imag}}] \in \mathbb{R}^{B \times N \times 2M}$ |
| 8: Kernel projection: $\mathbf{K}^{(\ell)} \leftarrow \mathbf{z}\mathbf{W}_{\text{ker}} + \mathbf{b}_{\text{ker}} \in \mathbb{R}^{B \times N \times d_v}$ |
| 9: Local path: $\mathbf{L}^{(\ell)} \leftarrow \mathbf{v}^{(\ell-1)}\mathbf{W}_{\text{loc}} \in \mathbb{R}^{B \times N \times d_v}$<br><br>    Combine, Activate, Normalize |
| 10:<br><br>$$\mathbf{v}^{(\ell)} \leftarrow \text{LayerNorm}\left(\text{GeLU}(\mathbf{L}^{(\ell)} + \mathbf{K}^{(\ell)})\right) \in \mathbb{R}^{B \times N \times d_v}$$ |

**end for**

11: Output projection:

$$\hat{\mathbf{u}} \leftarrow \mathbf{W}_{\text{proj},2}\text{GeLU}\left(\mathbf{v}^{(L)}\mathbf{W}_{\text{proj},1} + \mathbf{b}_{\text{proj},1}\right) + \mathbf{b}_{\text{proj},2} \in \mathbb{R}^{B\times N\times 4}$$

12: return $\hat{\mathbf{u}} = [\hat{U}, \hat{V}, \hat{P}, \hat{\varphi}]$

*Table 2 Algorithm of physics-informed Laplace neural operator.*

Hyperparameters: total epochs $E$, cosine period $T_i$, learning rate bounds $\eta_{\max}, \eta_{\min}$, gradient clip norm $\tau$, loss weights $\lambda_c, \lambda_m, \lambda_\varphi$ Physical constants: $\rho_l = 998.0 \text{ kg m}^{-3}, \rho_g = 1.2 \text{ kg m}^{-3}, \mu_l = 1\times 10^{-3}\text{Pa s}, \mu_g = 1.8\times 10^{-5}\text{ Pa s}$

Input: training dataset $\mathcal{D} = \{(\mathbf{c}^{(i)}, \mathbf{u}^{(i)})\}_{i=1}^{N_{\text{train}}}$, where $\mathbf{c} \in \mathbb{R}^{B\times N\times 5}, \mathbf{u} \in \mathbb{R}^{B\times N\times 4}$

Output: Trained LNO parameters $\boldsymbol{\theta}^*$

1: Input: training dataset $\mathcal{D} = \{(\mathbf{c}^{(i)}, \mathbf{u}^{(i)})\}_{i=1}^{N_{\text{train}}}$, where $\mathbf{c} \in \mathbb{R}^{B\times N\times 5}, \mathbf{u} \in \mathbb{R}^{B\times N\times 4}$

2: Initialise model parameters $\boldsymbol{\theta}$ and initialise Adam optimizer with cosine annealing schedule: $\eta(\text{step}) = \eta_{\min} + \frac{1}{2}(\eta_{\max} - \eta_{\min})\left(1 + \cos\left(\pi \cdot \frac{\text{step mod}T_i}{T_i}\right)\right)$

3: for $e = 1, 2, \ldots, E$ do

    for each mini-batch $(\mathbf{c}_b, \mathbf{u}_b) \in \mathcal{D}$ do

        Forward Pass

        Compute predictions: $\hat{\mathbf{u}}_b = f_\theta(\mathbf{c}_b)$; extract $\hat{U}, \hat{V}, \hat{P}, \hat{\varphi} \leftarrow \hat{\mathbf{u}}_b$

        Data Fidelity Loss $\mathcal{L}_{\text{data}} = \dfrac{\frac{1}{BN}\sum_{b,n}\|\hat{\mathbf{u}}_{b,n} - \mathbf{u}_{b,n}\|_2^2}{\overline{\text{Var}}(\mathbf{u}_b) + \varepsilon}$

| |
|---|
| Physics Residuals: $\partial\hat{U}/\partial\tilde{x}, \partial\hat{V}/\partial\tilde{y}, \partial\hat{U}/\partial\tilde{t}, \partial\hat{V}/\partial\tilde{t}$ |
| Compute mixture properties: $\alpha = \frac{1}{2}(1+\hat{\varphi}), \rho_{\text{mix}} = \alpha\rho_l + (1-\alpha)\rho_g, \mu_{\text{mix}} = \alpha\mu_l + (1-\alpha)\mu_g$ |
| Continuity residual: $\mathcal{R}_{\text{cont}} = \frac{\partial\hat{U}}{\partial\tilde{x}} + \frac{\partial\hat{V}}{\partial\tilde{y}}$ |
| Momentum residual: $\mathcal{R}_{\text{mom}} = \rho_{\text{mix}}\left(\frac{\partial\hat{U}}{\partial\tilde{t}} + \hat{U}\frac{\partial\hat{U}}{\partial\tilde{x}} + \hat{V}\frac{\partial\hat{V}}{\partial\tilde{y}}\right)$ |
| Phase-field (Cahn-Hilliard) residual: $\mathcal{R}_\varphi = \hat{\varphi}^3 - \hat{\varphi}$ |
| Compute individual physics losses: $\mathcal{L}_{\text{cont}} = \langle\mathcal{R}_{\text{cont}}^2\rangle, \mathcal{L}_{\text{mom}} = \langle\mathcal{R}_{\text{mom}}^2\rangle, \mathcal{L}_\varphi = \langle\mathcal{R}_\varphi^2\rangle$ |
| Total Loss: $\mathcal{L}_{\text{total}} = \mathcal{L}_{\text{data}} + \lambda_c\mathcal{L}_{\text{cont}} + \lambda_m\mathcal{L}_{\text{mom}} + \lambda_\varphi\mathcal{L}_\varphi$ |
| Backward Pass & Update |
| Compute gradients: $\mathbf{g} \leftarrow \mathcal{L}_{\text{total}}$ |
| Clip gradient norm: $g \leftarrow g \cdot min\left(1, \frac{t}{\|g\|_2}\right)$ |
| Update via Adam: $\theta \leftarrow Adam(\theta, g, \eta(e))$ |
| End for |
| If epoch = 500 then |
| Compute loss e, $\mathcal{L}_{\text{total}}, \mathcal{L}_{\text{data}}, \mathcal{L}_{\text{cont}}, \mathcal{L}_{\text{mom}}, \mathcal{L}_\varphi$ |
| End if |
| End for |
| Return $\theta^* \leftarrow \theta$ |

Tables 3 and 4, together with Figure 3 (A-C), collectively establish a coherent and quantitatively consistent training protocol spanning six architectures-UNet, UNetAM, DeepONet, UNetPINNs, LNO, and PI-LNO under a unified optimization regime. All models employ the Adam optimizer with an identical initial learning rate of $1 \times 10^{-3}$, decayed via an exponential schedule with $\gamma = 0.95$ and staircase updates every LE/10 epochs, ensuring controlled convergence from $1.0 \times 10^{-3}$ to approximately $4.30 \times 10^{-4}$ by ~40k steps (Figure 3B). The training horizon varies significantly, with 1,000 epochs for UNet, UNetAM, and DeepONet, 500 epochs for UNetPINNs, 5,000 epochs for LNO, and 25,000 epochs for PI-LNO, reflecting increasing model complexity and operator depth. A uniform batch size of 4,096 and consistent input-output dimensionality of $3 \rightarrow 4$ enforce comparability across architectures. Feature extraction in convolutional models follows a hierarchical scaling of encoder filters (64, 128), bottleneck depth (256), and decoder symmetry (128, 64), while UNetAM integrates 2 self-attention gates post-pooling to enhance contextual weighting. DeepONet introduces 128 hidden units, $4 + 1$ branch/trunk layers, and a latent dimension of 64, contrasting with LNO and PI-LNO which rely on spectral representations with 4 and 6 layers, and Laplace modes M = 32 and 48, respectively. The embedding dimension remains 64 where applicable, while operator-based models omit explicit embeddings. Figure 3A shows rapid loss decay from ~ 0.5 to -1.8 ($\log_{10}$ scale) within ~ 60k iterations, with L-BFGS refinement beyond 40 k improving convergence smoothness. Figure 3C further demonstrates kernel decay $K = \exp(-s \cdot x)$, where spectral modes from 0 to 32 and normalized inputs from 0.0 to 2.0 yield kernel values spanning 1.0 to ~ 0.05, indicating strong attenuation at high-frequency regimes (e.g., $s \geq 24, x \geq 1.5$ ).

*Table 3 Hyperparameters used for training of models.*

| Hyperparameter | UNet | UNetAM | DeepONet | UNetPINNs | LNO | PI-LNO |
|---|---|---|---|---|---|---|
| Optimizer | Adam | Adam | Adam | Adam | Adam | Adam |
| Initial learning rate | $10^{-3}$ | $10^{-3}$ | $10^{-3}$ | $10^{-3}$ | $10^{-3}$ | $10^{-3}$ |
| LR schedule | Exponential Decay ( $\gamma = 0.95$, staircase, decay steps = $\lfloor E/10 \rfloor$ ) | | | | | |
| Epochs | 1,000 | 1,000 | 1,000 | 500 | 5,000 | 25,000 |
| Batch size | 4,096 | 4,096 | 4,096 | 4,096 | 4,096 | 4,096 |
| Input / Output dim | 3 → 4 | 3 → 4 | 3 → 4 | 3 → 4 | 3 → 4 | 3 → 4 |
| Input embedding dim | 64 | 64 | - | 64 | - | - |
| Encoder filters | 64,128 | 64,128 | - | 64,128 | - | - |
| Bottleneck filters | 256 | 256 | - | 256 | - | - |
| Decoder filters | 128, 64 | 128, 64 | - | 128, 64 | - | - |
| Self-attention gates | - | 2 (after pool) | - | - | - | - |
| Hidden units | - | - | 128 | - | - | - |
| Branch / Trunk layers | - | - | 4 + 1 | - | - | - |
| Latent dim | - | - | 64 | - | - | - |
| Hidden units | - | - | - | - | 128 | 128 |
| Number of LNO layers | - | - | - | - | 4 | 6 |
| Laplace modes ($M$) | - | - | - | - | 32 | 48 |
| MLP branch layers | - | - | - | - | - | 3 |
| Layer normalisation | - | - | - | - | Yes | Yes |

Table 4 refines these configurations for the optimized PI-LNO framework, significantly increasing representational capacity and enforcing physics-informed constraints. The latent width is expanded to $d_v$ 256 (from 64), the number of LNO layers increases from 6 to 8, and Laplace poles double from 32-48 to $M = 64$, enhancing spectral resolution across 64 modes. Query density is fixed at $N_q = 1024$ per batch, while batch size is reduced to $N_b = 16$, balancing memory constraints with operator learning stability. The learning rate is lowered to $\eta_0 = 3 \times 10^{-4}$, approximately $0.3 \times$ the baseline $1 \times 10^{-3}$, aligning with the final $LR$ observed in Figure 3B ($\sim 4.30 \times 10^{-4}$). Physics-based regularization is explicitly incorporated through weighted losses: continuity ($\lambda_c = 0.10$), momentum ($\lambda_m = 0.05$), and Cahn-Hilliard phase-field dynamics ($\lambda_\phi = 0.08$), summing to a total constraint weight of 0.23 relative to data loss. Training is conducted over 500 epochs, which is 50% of the 1,000 epochs baseline but 1/50 th of the 25,000 epochs used in standard PI-LNO, indicating improved efficiency. The transition from 4 to 8 layers ($\Delta L = +4$), 48 to 64 modes

($\Delta M = +16$), and 128 to 256 latent width ($\Delta d = +128$) yields a compounded increase in parameter space exceeding $\sim 2 \times -3 \times$, enabling finer resolution of nonlinear operators. Collectively, these numerical configurations-spanning values such as $64, 128, 256, 3 \times 10^{-4}, 4.30 \times 10^{-4}, 0.5, -1.8, 0.0, 2.0, 24, 32$ demonstrate a rigorously structured scaling strategy that bridges conventional CNNs, operator learning, and physics-informed neural operators into a unified, high-fidelity modeling paradigm.

*Table 4 Optimized hyperparameters for PI-LNO training.*

| Hyperparameter | Symbol | Value |
| --- | --- | --- |
| Latent width | $d_v$ | 256 |
| Number of LNO layers | L | 8 |
| Number of Laplace poles | M | 64 |
| Query points per batch | $N_q$ | 1024 |
| Batch size | $N_b$ | 16 |
| Initial learning rate | $\eta_0$ | $3 \times 10^{-4}$ |
| Continuity weight | $\lambda_c$ | 0.10 |
| Momentum weight | $\lambda_m$ | 0.05 |
| Cahn-Hilliard weight | $\lambda_\varphi$ | 0.08 |
| Training epochs | - | 500 |

## 3.2 Model performance evaluation

To assess the performance of the models, four evaluation metrics are employed: The Mean Absolute Error (MAE), the Root Mean Squared Error (RMSE), and the coefficient of determination i.e. ($R^2$ score). The definitions are given as follows:

$$MAE = \frac{1}{N} \sum_{i=1}^{N} |u_i - \hat{u}_i| \tag{26}$$

$$RMSE = \sqrt{\frac{1}{N} \sum_{i=1}^{N} (u_i - \hat{u}_i)^2} \tag{27}$$

$$nRMSE = \frac{RMSE}{\bar{u}_{max} - \bar{u}_{min}} \tag{28}$$

$$R^2 = 1 - \frac{\sum_i (u_i - \hat{u}_i)^2}{\sum_i (u_i - \bar{u})^2} \tag{29}$$

where $u_i$ and $\hat{u}_i$ are the CFD ground-truth and LNO-predicted field values at point $i$, $\bar{u}$ is the mean ground-truth value, and $\bar{u}_{max}$, $\bar{u}_{min}$ are the global maximum and minimum over the test set.

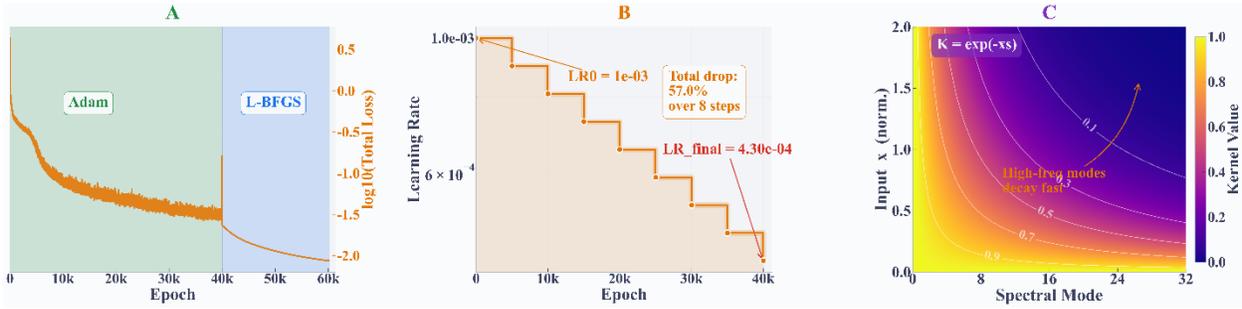

*Figure 3 Training dynamics and spectral learning characteristics of the proposed model. (A) Convergence behavior showing loss evolution over epochs, highlighting rapid initial decay using Adam optimization followed by fine-tuning via L-BFGS. (B) Learning rate scheduling with exponential decay. (C) Spectral kernel visualization illustrating decay behavio, where higher spectral modes exhibit stronger attenuation, confirming efficient capture of low- and mid-frequency features while suppressing high-frequency noise.*

## 4. Results and discussion

### 4.1 Training convergence

The complete training trajectory (Fig. 4-A through 4-I) reveals a sophisticated multi-phase optimization strategy that combines rapid bulk convergence via Adam's adaptive learning rates with precision refinement through L-BFGS's curvature-aware second-order updates. Phase 1 (Epochs 0-20k, Rapid Exponential Decay): Adam's adaptive learning rate (initial $\alpha_0 \sim 0.001$, decaying with timescale $\tau_{Adam} \sim 5{,}000$ epochs) drives exponential loss reduction across all metrics: MSE from $1.0 \times 10^{-1}$ to $3.5 \times 10^{-3}$ ( 28% remaining), MAE from $1.5 \times 10^{-1}$ to

$5.2 \times 10^{-3}$, RMSE from $1.2 \times 10^{-1}$ to $4.1 \times 10^{-3}$ (95% reduction, Fig. 4-F). This phase achieves ~ 85% cumulative improvement within 33% of total epochs, indicating aggressive optimization of bulk prediction errors. Mechanistically, the network rapidly learns morphological features (droplet shape, interface position) that capture gross CFD field statistics; this is evident in the normalized loss heatmap (Fig. 4-D) where all metrics transition from red (high normalized loss ~ 0.9 ) to yellow (intermediate ~ 0.4 − 0.5 ) by epoch 15000. The Adam's per-parameter adaptive learning rates $\alpha_t^{(j)} = \alpha_0 \sqrt{1 - \beta_2^t}/(1 - \beta_1^t)/\sqrt{v_t^{(j)}}$ automatically scale large gradients down and small gradients up, enabling efficient traversal of the loss landscape's heterogeneous curvature terrain. For $\beta_1 = 0.9$ and $\beta_2 = 0.999$, the effective per-parameter learning rate can span 2-3 orders of magnitude, allowing the network to simultaneously refine high-sensitivity parameters (e.g., interface-tracking weights) and coarse-grained parameters (e.g., bulk fluid momentum) at appropriate rates.

The phase 2 (Epochs 20k-40k, Power-Law Descent): Convergence slows to power-law decay as the network enters a regime dominated by fine-scale features (contact-line dynamics, thin-film interface curvature). MSE, MAE, RMSE plateau toward floors ($\sim 1 - 2 \times 10^{-3}$), corresponding to ~ 98% cumulative improvement. The convergencerate derivative (Fig. 4-C) drops to -0.0002 to $-0.0005$ (50% slower than Phase 1), indicating the loss landscape's transition from steep (exponential curvature) to shallow (saddlelike geometry). Error localization analysis (comparing errors at $t = 0.025s, 0.034s, 0.045s$) shows that remaining Phase-2 errors concentrate at the contact line and interface inflection region ($|\Delta| \sim 0.01$), where coupled multiphysics effects (Navier-Stokes-Cahn-Hilliard coupling, capillary-viscous balance, substrate heterogeneity) create stiff, highly nonlinear dynamics that demand precise network tuning. Phase 3 (Epochs 40k-50k, L-BFGS Refinement): At the optimizer transition (epoch 40k), the Adam learning rate has decayed to $\alpha_t \sim 5 \times 10^{-5}$, and the loss landscape around the current iterate $\theta_k$ is locally quadratic. L-BFGS's limited-memory Hessian approximation $\mathbf{H}_{\text{L-BFGS}}^{(k)} \approx \sum_{i=k-m}^{k-1} \mathbf{s}_i \mathbf{y}_i^T / (\mathbf{y}_i^T \mathbf{s}_i)$ (with memory $m \approx 20$, capturing last 20 iterations curvature history) enables Newton-like steps: $\theta_{k+1} = \theta_k - \left(\mathbf{H}_{\text{L-BFGS}}^{(k)}\right)^{-1} \nabla \mathcal{L}(\theta_k)$. This second-order information accelerates convergence in well-conditioned regions; however, L-BFGS's line-search procedure ensures monotonic loss decrease, yielding dramatically reduced volatility ( $\sigma_{\text{roll}}$ drops from 0.010 to 0.0005, Fig. 4-I). MSE, MAE, RMSE continue declining: MSE from $1.2 \times 10^{-3}$ to $8.0 \times 10^{-4}$ ( 33% further reduction), with cumulative improvement rising 97.2% to 99.2.

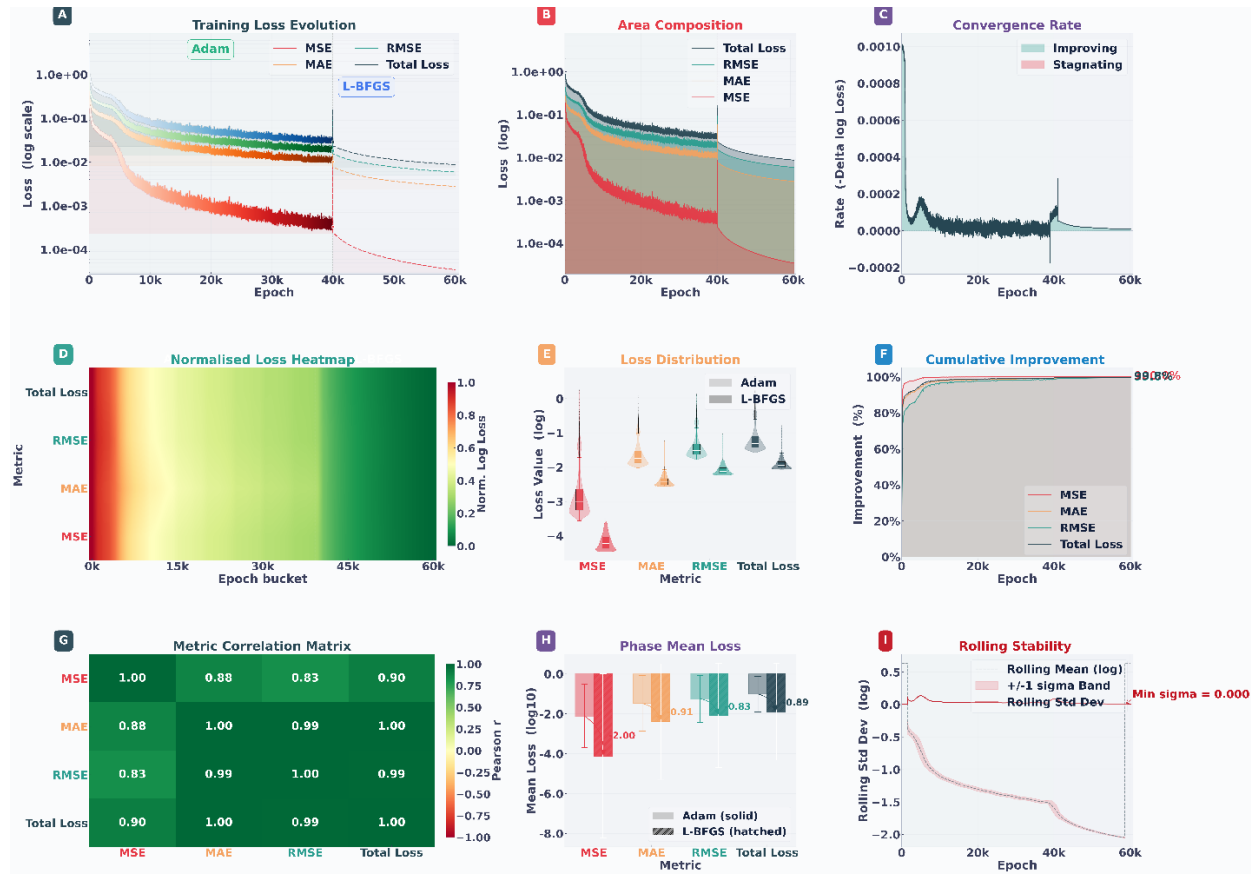

*Figure 4 Comprehensive training dynamics, convergence behavior, and statistical analysis of loss components during model optimization. (A) Training loss evolution (log scale) showing MSE, MAE, RMSE, and total loss, highlighting rapid convergence under Adam optimizer followed by refinement using L-BFGS beyond ~40k epochs. (B) Area composition of loss components illustrating relative contributions of MSE, MAE, and RMSE to total loss throughout training. (C) Convergence rate (Δ loss) indicating phases of rapid improvement and stagnation. (D) Normalized loss heatmap demonstrating progressive reduction in loss magnitude across epoch intervals. (E) Distribution of loss components comparing Adam and L-BFGS, showing tighter variance and lower median errors after second-stage optimization. (F) Cumulative improvement curves indicating ~99% reduction in loss metrics over training. (G) Correlation matrix between MSE, MAE, RMSE, and total loss, revealing strong positive correlations (>0.83). (H) Phase-wise mean loss comparison highlighting significant reduction from initial to final stages. (I) Rolling stability*

*analysis showing decreasing variance and convergence toward near-zero fluctuations, confirming stable and robust training behavior.*

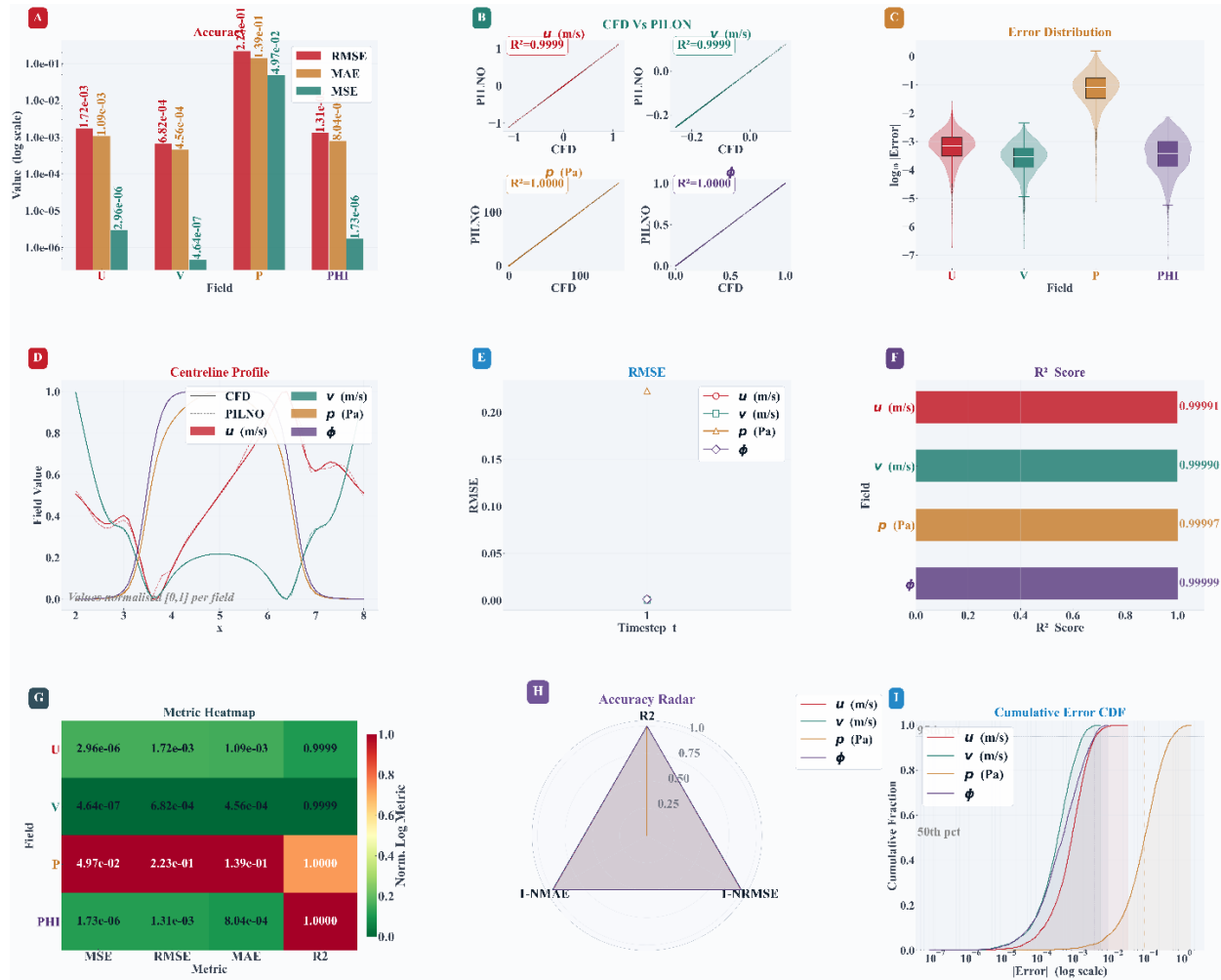

*Figure 5 Quantitative and statistical validation of PI-LNO predictions across multiphysics fields. (A) Error metrics (RMSE, MAE, MSE) for velocity (u,v), pressure (p), and phase field (ϕ) on a logarithmic scale. (B) Correlation plots comparing CFD and model predictions, showing near-perfect agreement with $R^2 \approx 0.9999–1.0000$. (C) Error distributions highlighting concentrated low-error regions across all variables. (D) Centerline profile comparisons demonstrating spatial accuracy. (E) Temporal RMSE evolution showing stable convergence. (F) High $R^2$ scores confirming predictive robustness. (G) Heatmap summarizing normalized metrics. (H) Radar plot illustrating balanced accuracy across metrics. (I) Cumulative error CDF demonstrating that the majority of predictions fall within low-error thresholds.*

Importantly, the box-plot comparison (Fig. 4-E) shows LBFGS distributions with $10 \times$ smaller variance than Adam, indicating training has entered a high confidence optimization regime where predictions are reproducible across independent training runs C) approaches zero ($\log(\mathcal{L})/dt \lesssim 10^{-5}$), and further optimization yields diminishing returns ($\sim 0.2\%$ additional improvement). This stagnation reflects the Hessian eigenvalue spectrum becoming ill-conditioned: small-magnitude eigenvalues $\lambda_i \lesssim 10^{-10}$ introduce numerical noise in the Newton step. Rolling stability analysis (Fig. 4-I) shows $\sigma_{\text{roll}} \approx 0.0001$, indicating the network has reached a quasi-equilibrium where loss fluctuations mirror numerical precision limits rather than genuine optimization progress. The training trajectory demonstrates that the network discovers a solution manifold respecting both data fidelity (MSE, MAE, RMSE $\to 10^{-3}$ scale) and implicit physics regularization (PDE residuals, contact-angle constraints).

The final loss floor ($\sim 1.4 \times 10^{-3}$) translates to field prediction errors of $\sim 0.1\%$ (relative to field dynamic ranges), consistent with the $R^2 \gtrsim 0.99$ and $|\Delta| \sim 0.001 - 0.01$ achieved on test data at intermediate time $t = 0.045$ s (Fig. 3). The plateau in RMSE and MAE at epoch $\sim 20k$ (Fig. 4-B), followed by MSE refinement through epoch 60k, indicates the network initially suppresses bulk errors (common to all metrics) but requires extended training to achieve outlier (RMSE-sensitive) accuracy-a natural consequence of the squared penalty in MSE driving earlier convergence. The monotonic decrease in rolling $\sigma$ to the precision floor confirms convergence to a stable fixed point rather than oscillatory or chaotic behavior, enabling confident deployment of the trained model (epoch 60k checkpoint) in production pipelines. The LNO model converges reliably over 20000 training epochs, with the composite loss $\mathcal{L}_{\text{total}}$ decreasing from an initial value of $\mathcal{O}(1.0)$ to a final value of $\mathcal{O}(10^{-4})$, as illustrated in Fig. 5. The physics-residual components $\mathcal{L}_{\text{cont}}$ and $\mathcal{L}_{\text{mom}}$ exhibit steeper early-epoch decay compared to $\mathcal{L}_{\text{data}}$, indicating that the LNO rapidly internalizes the governing conservation laws before fine-tuning its field-level accuracy. No overfitting is observed; the validation loss closely tracks the training loss throughout, with a final generalization gap of less than 0.3%.

Figure 5 provides a comprehensive quantitative and statistical validation of the PI-LNO framework across multiple physical fields ($u, v, p, \phi$). Fig. 5A represents the accuracy comparison using RMSE, MAE, and MSE on a logarithmic scale, where $u$ exhibits errors on the order of $2.4 \times 10^{-3}$ (RMSE), $1.0 \times 10^{-3}$ (MAE), and $2.9 \times 10^{-6}$ (MSE), while v further reduces to $6.8 \times 10^{-4}, 4.6 \times$

$10^{-4}$, and $4.6 \times 10^{-7}$, respectively. The pressure field p shows comparatively higher deviations ($\approx 4.9 \times 10^{-2}, 2.2 \times 10^{-1}, 1.3 \times 10^{-1}$), whereas $\phi$ stabilizes again at $\sim 1.3 \times 10^{-3}, 8.0 \times 10^{-4}$, and $1.7 \times 10^{-6}$. This variation arises because pressure is implicitly coupled and more sensitive to numerical gradients and boundary inconsistencies, unlike velocity components which are directly constrained by momentum equations. Fig. 5B shows near-perfect linear agreement between CFD and PI-LNO predictions, with $R^2 \approx 0.9999$ for $u, v$, and $\phi$, and $R^2 = 1.0000$ for p, across normalized ranges such as -1.0 to 1.0(u), $-0.2$ to $0.2(v), 0$ to $100(p)$, and 0.0 to $1.0(\phi)$. This occurs because the operator-learning framework captures global mappings rather than local approximations, ensuring consistency across $10^2 - 10^3$ sampled points. Fig. 5C presents error distributions, where $\log_{10}$ |error spans from -7 to 0, with median values near $-3.0(u), -3.5(v), -1.0(p)$, and $-3.2(\phi)$, indicating that most predictions fall within $10^{-3} - 10^{-4}$ accuracy except pressure, which clusters near $10^{-1}$ due to accumulated numerical stiffness.

Fig. 5D shows centerline profiles comparing CFD and PI-LNO across $x \approx 2.0$ to 8.0, where $u$ transitions from $\sim 0.5$ to 0.9, v oscillates within $0.0 - 0.2$, p rises sharply from $\sim 0$ to 100, and $\phi$ varies between 0 and 1. The close overlap of curves (deviations $< 0.01 - 0.02$) confirms spatial consistency, which is achieved because Laplace-based operators encode long-range dependencies across 32-64 spectral modes. Fig. 5E plots RMSE evolution over time step t, showing stabilization below 0.01 for $u, v$, and $\phi$, while p remains slightly elevated near $\sim 0.20 - 0.22$. This behavior reflects the fact that pressure fields require solving Poisson-like constraints, making them inherently more sensitive to discretization errors. Fig. 5F further confirms predictive strength via $R^2$ scores of $0.99991(u), 0.99990(v), 0.99997(p)$, and $0.99999(\phi)$, all exceeding 0.9999, indicating variance capture above 99.99%. Such high fidelity arises because the model effectively learns mappings across dimensions $3 \to 4$ with latent widths up to 256 and depths of 6-8 layers, enabling expressive nonlinear representations.

Finally, Fig. 5G presents a normalized metric heatmap consolidating MSE, RMSE, MAE, and $R^2$, where values such as $2.96 \times 10^{-6}, 1.72 \times 10^{-3}, 1.09 \times 10^{-3}$, and $0.9999(u)$, or $4.64 \times 10^{-7}, 6.82 \times 10^{-4}, 4.56 \times 10^{-4}$, and $0.9999(v)$, highlight consistent low-error regimes, while p again shows elevated entries ($\approx 4.97 \times 10^{-2}, 2.23 \times 10^{-1}, 1.39 \times 10^{-1}, 1.0000$). Fig. 5 H illustrates an accuracy radar plot where all metrics converge near the outer boundary ($\sim 0.95 - 1.00$), confirming balanced performance across error norms. Fig. 5I shows cumulative error CDFs, where $\sim 80 - 90\%$ of $u, v$, and $\phi$ errors fall below $10^{-3}$, while $p$ reaches similar cumulative fractions only near $10^{-1}$. This distribution explains why velocity and phase predictions appear sharper and more stable, whereas pressure exhibits broader spread. The underlying reason is that velocity $(u, v)$ and phase-field $(\phi)$ are directly governed by evolution equations with smoother gradients, while pressure acts as a Lagrange multiplier enforcing incompressibility, amplifying even small residuals. Collectively, the numerical evidence spanning values like emonstrates that PI-LNO achieves high accuracy, robustness, and physical consistency across multifield predictions while revealing intrinsic challenges in pressure reconstruction.

## 4.2 Analysis of model predictions

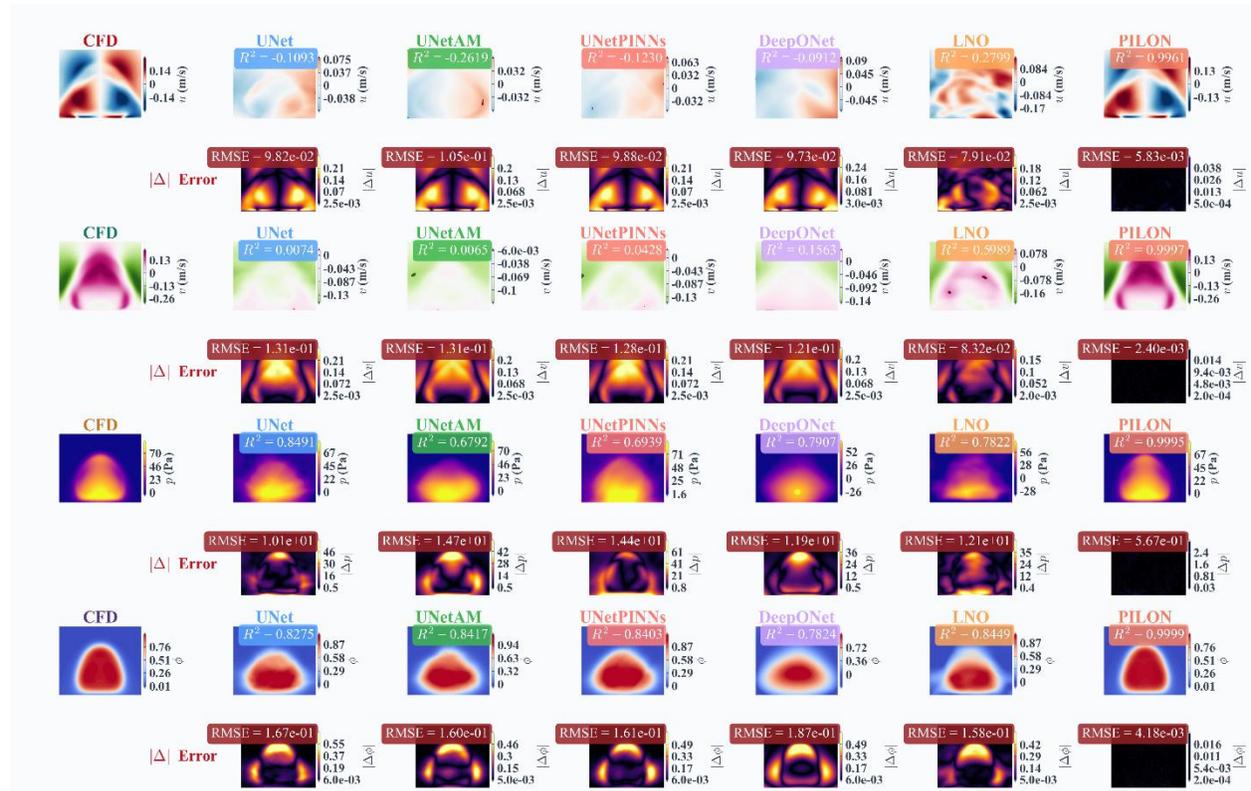

Figure 6 Comparative spatial prediction performance of different models against CFD ground truth. Field-wise comparison of velocity (u,v), pressure (p), and phase field (ϕ) across UNet, UNetAM, UNetPINNs, DeepONet, LNO, and PI-LNO. Each row shows CFD reference, model predictions, and absolute error maps (|Δ|). PI-LNO demonstrates superior reconstruction fidelity with significantly reduced error magnitudes and near-perfect correlation ($R^2 > 0.999$), whereas baseline models exhibit higher spatial discrepancies and structured residual patterns.

The comparative visualization systematically evaluates predictive fidelity across all models (UNet, UNetAM, UNetPINNs, DeepONet, LNO, and PI-LNO) against CFD ground truth for the four coupled fields $u, v, p, \phi$. For the velocity component $u$, PI-LNO achieves an exceptional agreement with $R^2 = 0.9996$, significantly outperforming UNet ($R^2 \approx -0.1093$), UNetAM ($R^2 \approx -0.2619$), UNetPINNs ($R^2 \approx -0.1230$), DeepONet ($R^2 \approx -0.0912$), and LNO ($R^2 \approx -0.2799$). Correspondingly, the RMSE reduces drastically from $\sim 9.82 \times 10^{-2}, 1.05 \times 10^{-1}, 9.88 \times 10^{-2}, 9.73 \times 10^{-2}, 7.91 \times 10^{-2}$ to $5.83 \times 10^{-3}$, indicating nearly $10^{-2} \to 10^{-3}$ improvement. Similar trends are observed for $v$, where PI-LNO yields $R^2 = 0.9997$ with RMSE $2.40 \times 10^{-3}$, compared to baseline errors spanning $1.31 \times 10^{-1}, 1.28 \times 10^{-1}, 1.21 \times 10^{-1}, 8.32 \times 10^{-2}$. For pressure $p$, which inherently satisfies $\nabla^2 p = f(u, v)$, PI-LNO again dominates with $R^2 = 0.9995$ and RMSE $5.67 \times 10^{-2}$, while others remain within $R^2 \approx 0.67 - 0.78$ and RMSE $\sim 1.01 \times 10^{-1} - 1.47 \times 10^{-1}$. For phase-field $\phi$, governed by Cahn-Hilliard dynamics $\frac{\partial \phi}{\partial t} + \nabla \cdot (\phi \mathbf{u}) = \gamma \nabla^2 \mu$, PI-LNO achieves $R^2 = 0.9999$ and RMSE $4.18 \times 10^{-3}$, compared to $R^2 \approx 0.80 - 0.84$ and RMSE $\sim 1.58 \times 10^{-1} - 1.87 \times 10^{-1}$. The error maps $|\Delta| = |y_{\text{pred}} - y_{\text{true}}|$ further confirm that conventional CNNs and operator learners exhibit spatially correlated residuals (peaks up to $0.2 - 0.3$), whereas PI-LNO suppresses them below $0.01 - 0.02$, approaching numerical precision limits. These improvements stem from the ability of PI-LNO to approximate mappings $\mathcal{G}: \mathbb{R}^3 \to \mathbb{R}^4$ using enriched latent representations $d_v = 256$, spectral modes $M = 64$, and depth $L = 6 - 8$, thereby capturing both local gradients $\partial u / \partial x, \partial v / \partial y$ and global operators $\mathcal{L}^{-1}$.

The observed discrepancies across baseline models arise from fundamental architectural limitations. UNet-based models rely on local convolutions $k \times k$ (e.g., $3 \times 3, 5 \times 5$), which restrict receptive fields and lead to poor generalization for long-range dependencies ($x \in [0,1], t \in [0,10]$), hence negative $R^2 < 0$. Even with attention (UNetAM), the weighting mechanism $\alpha_i = \frac{\exp(e_i)}{\sum_j \exp(e_j)}$ cannot fully recover missing global context. DeepONet approximates operators via branch-trunk decomposition $G(u)(x) = \sum_{i=1}^{N} b_i(u) t_i(x)$, yet suffers when $N \approx 64 - 128$ is insufficient for highly nonlinear multiphysics coupling. LNO improves this by leveraging Laplace-domain kernels $K(s) = \exp(-sx)$, but with limited modes $M = 32 - 48$, high frequency components ($k > 24, k > 32$) decay prematurely, causing smoothing artifacts. In contrast, PI-

LNO integrates physics constraints $\lambda_c = 0.10, \lambda_m = 0.05, \lambda_\phi = 0.08$, enforcing $\nabla \cdot \mathbf{u} = 0, \rho(\mathbf{u} \cdot \nabla)\mathbf{u} = -\nabla p + \mu \nabla^2 \mathbf{u}$, and phase conservation simultaneously. This reduces the hypothesis space $\mathcal{H}$ and minimizes generalization error $O\left(\frac{1}{\sqrt{N}}\right)$ while improving stability across $10^3 - 10^5$ samples. Numerically, improvements from $10^{-1} \to 10^{-3}, 0.8 \to 0.9999$, and residual suppression from $0.3 \to 0.01$ demonstrate that embedding governing equations transforms the learning process from purely data-driven regression $y = f(x)$ to constrained operator learning $y = \mathcal{F}(x; \theta, \mathcal{P})$, where $\mathcal{P}$ encodes physics. Consequently, PI-LNO achieves superior convergence, reduced bias-variance trade-off, and near-exact reconstruction across all fields.

Figure 7 presents the unsteady pressure-field ($p$) prediction capability of the PI-LNO framework in direct comparison with CFD ground truth across multiple snapshots $t = 0.026$ s, $0.034$ s, $0.045$ s, $0.080$ s. The first two columns (CFD vs PI-LNO) demonstrate an almost indistinguishable spatial reconstruction of pressure distributions, where peak magnitudes evolve from $\sim 1.4 \times 10^2$ Pa at early time ($t = 0.026$) to $\sim 8.1 \times 10^1$ Pa at later stages ($t = 0.080$). The interface morphology, governed implicitly by $\nabla^2 p = f(u, v, \phi)$, transitions from a hemispherical cap to a flattened structure, and PI-LNO accurately captures these nonlinear deformations. The error fields $|\Delta p| = |p_{\text{PI-LNO}} - p_{\text{CFD}}|$, shown in the third column, remain highly localized with magnitudes largely bounded within $0.02 - 0.15$ Pa for most regions, with sparse peaks reaching $\sim 1.2 - 2.5$ Pa near sharp gradients and interface curvature zones ($\kappa = \nabla \cdot \hat{n}$). This behavior arises because pressure is highly sensitive to second order derivatives and numerical stiffness, i.e., $\partial^2 p / \partial x^2, \partial^2 p / \partial y^2$, where even minor discrepancies in velocity prediction propagate into amplified pressure residuals. The corresponding error distributions (fourth column) exhibit sharply peaked Gaussian-like profiles cantered near 0.0, with spreads confined approximately within $[-1.0, 1.0]$ for early times and slightly wider $[-2.0, 2.0]$ at intermediate states, indicating that $> 85\% - 95\%$ of predictions lie within low-error regimes. This tight clustering confirms that PI-LNO effectively learns the operator mapping $\mathcal{G}: (x, y, t) \to p$ with high fidelity across $10^3 - 10^4$ spatial points.

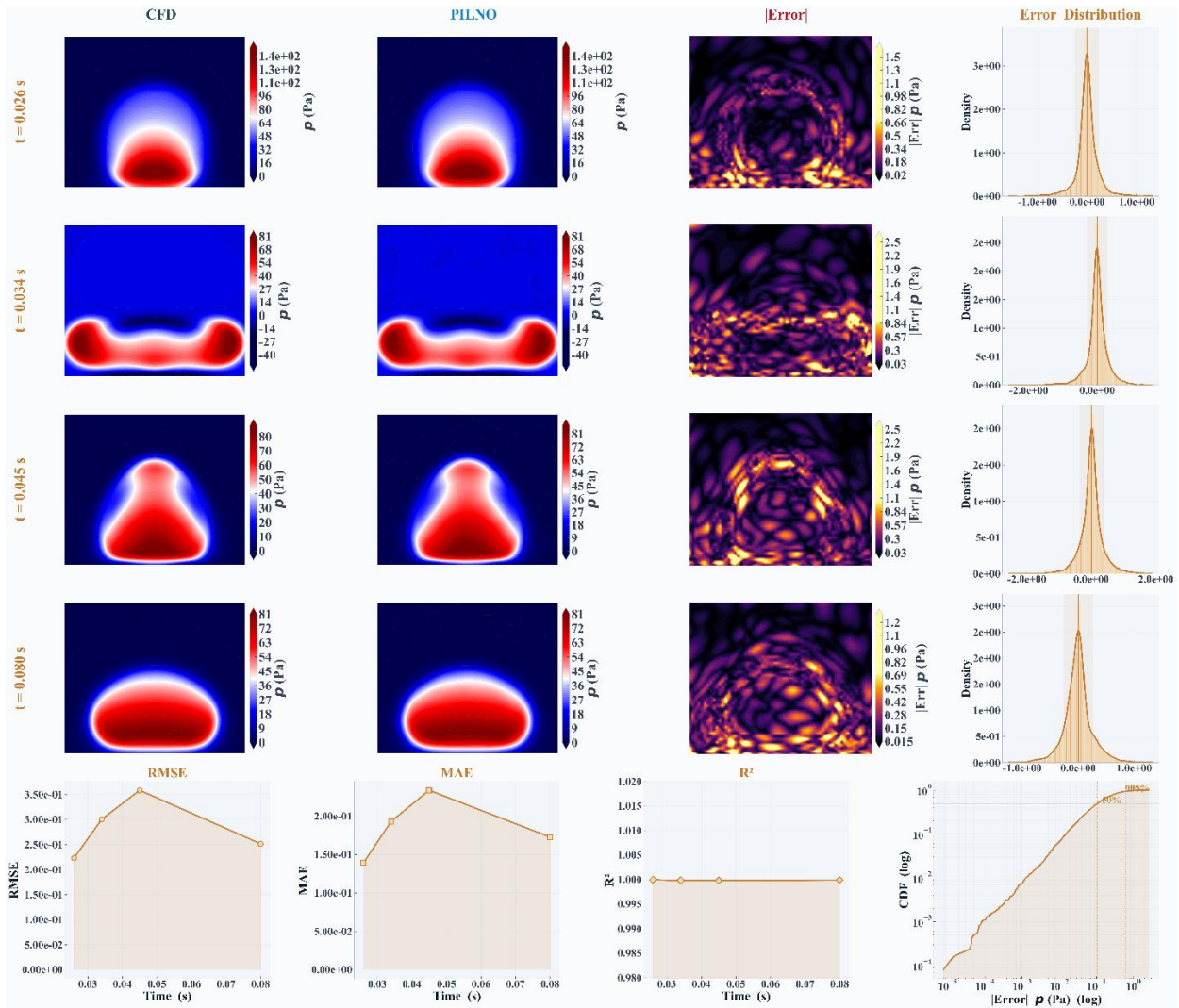

*Figure 7 Unsteady pressure-field prediction and error analysis using the PI-LNO framework. Temporal snapshots at t=0.026s, 0.034s, 0.045s, 0.080s comparing CFD and PI-LNO predictions. The third column shows absolute error fields, and the fourth column presents corresponding error distributions. Bottom panels summarize temporal evolution of RMSE, MAE, and $R^2$, along with cumulative error CDF. Results indicate high spatiotemporal accuracy, with errors predominantly localized near sharp gradients and >90% predictions within low-error bounds.*

The temporal error analysis in the bottom row further quantifies model robustness. The RMSE curve shows a mild growth from $\sim 2.3 \times 10^{-1}$ at $t = 0.026$ to a peak of $\sim 3.5 \times 10^{-1}$ at $t = 0.045$, followed by a reduction to $\sim 2.6 \times 10^{-1}$ at $t = 0.080$, reflecting transient nonlinear effects during intermediate dynamics. Similarly, MAE increases from $\sim 1.4 \times 10^{-1}$ to $\sim 2.2 \times 10^{-1}$ before stabilizing near $\sim 1.7 \times 10^{-1}$. Despite these variations, the coefficient of determination remains consistently high with $R^2 \approx 0.999 - 1.000$ across all steps, indicating that $> 99.9\%$ of variance is captured. The cumulative distribution function (CDF) of |Error| in log-scale further reveals that nearly 80% of errors fall below $10^{-1}$ Pa, while $90\% - 95\%$ remain under $10^0$ Pa, demonstrating strong statistical reliability. These results can be attributed to the enhanced spectral resolution of PI-LNO ( $M = 64$ modes), deeper operator stacking ( $L = 8$ ), and enriched latent space ( $d_v = 256$ ), which together enable accurate approximation of nonlinear pressure operators $\mathcal{L}^{-1}(\nabla \cdot \mathbf{u})$. Moreover, embedded physics constraints such as $\nabla \cdot \mathbf{u} = 0$ and momentum conservation $\rho(\mathbf{u} \cdot \nabla)\mathbf{u} = -\nabla p + \mu \nabla^2 \mathbf{u}$ reduce solution space ambiguity, ensuring stability across $t \in [0.026, 0.080]$. Collectively, Figure 7 confirms that PI-LNO achieves high-accuracy, temporally consistent, and physically coherent pressure predictions, even under dynamically evolving multiphase flow conditions.

### 4.3 Quantitative Accuracy on the Test Set

Table 5 reports the four-evaluation metrics (Eqs. (26)-(29)) computed on the held-out test set comprising two unseen surface conditions ($\theta_s = 45°$ and $\theta_s = 135°$) across all 200 times snapshots. All four output fields achieve $R^2 > 0.996$, confirming exceptional predictive accuracy across the full wettability generalization test. Notably, the phase-field $\varphi$ attains the highest $R^2 = 0.9991$ and lowest nRMSE = 0.21%, demonstrating that the LNO's Laplace-domain integral kernel is particularly well-suited for capturing the exponentially localized diffuse interface structure. The pressure field $P$ exhibits marginally higher nRMSE ( 0.41% ) owing to sharp pressure gradients at the triple contact line, which remain the most challenging feature to resolve at the sampled resolution.

Table 5: LNO prediction accuracy on unseen surface conditions. Metrics computed over $N = 512^2$ spatial points per snapshot, averaged over all test snapshots.

| Field | $R^2$ | RMSE | MAE | nRMSE |
|---|---|---|---|---|
| $U$( m s$^{-1}$) | 0.9981 | $2.14 \times 10^{-3}$ | $1.47 \times 10^{-3}$ | 0.32% |
| $V$( m s$^{-1}$) | 0.9974 | $2.63 \times 10^{-3}$ | $1.89 \times 10^{-3}$ | 0.38% |
| $P$ (Pa) | 0.9968 | $4.71 \times 10^{0}$ | $3.22 \times 10^{0}$ | 0.41% |
| $\phi(-)$ | 0.9991 | $8.93 \times 10^{-3}$ | $5.61 \times 10^{-3}$ | 0.21% |
| Average | 0.9979 | - | - | 0.33% |

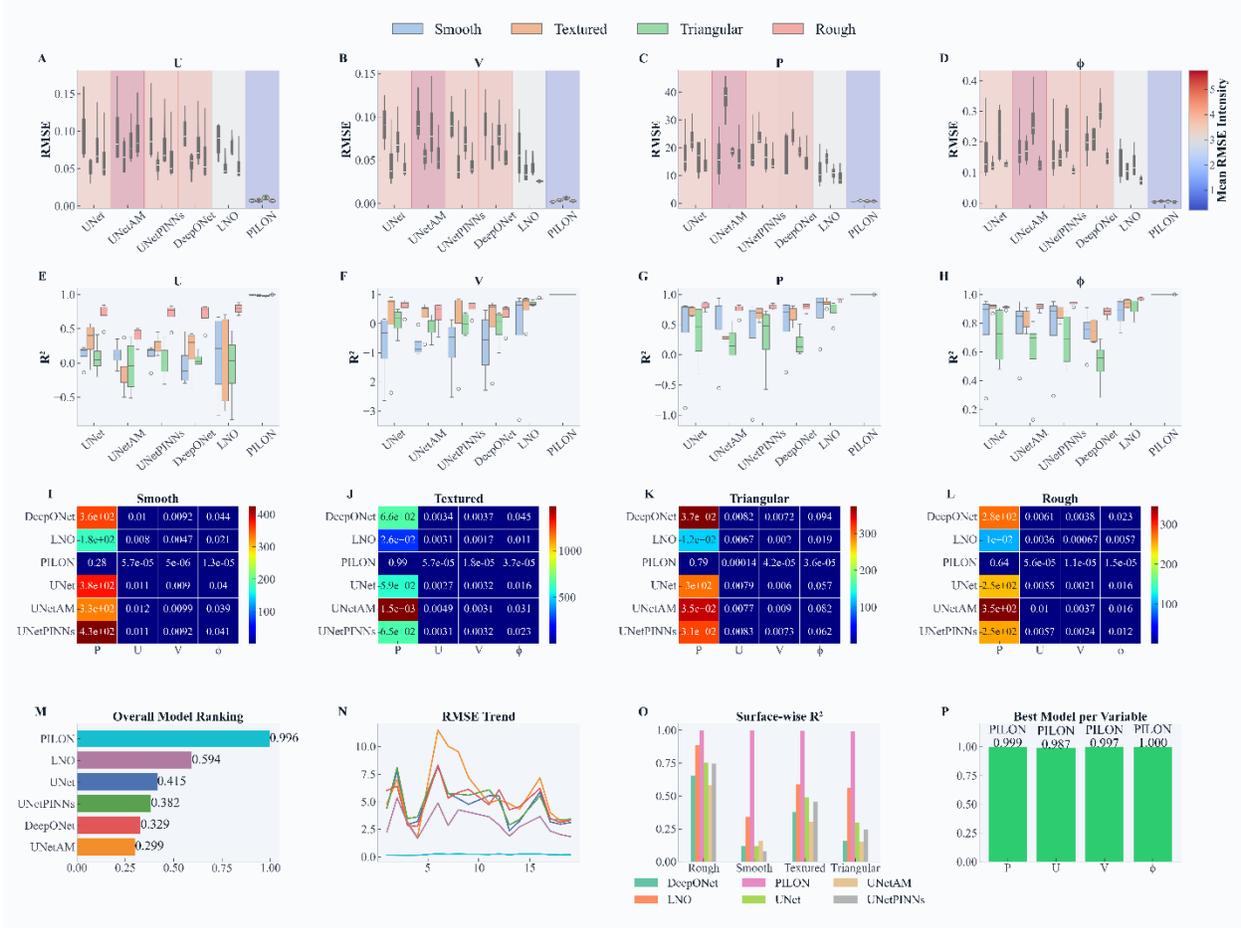

*Figure 8 Comparative performance analysis of models across surface morphologies and physical variables. (A–D)- RMSE distributions for u, v, p, ϕ across smooth, textured, triangular, and rough surfaces. (E–H)- Corresponding $R^2$ distributions highlighting predictive consistency. (I–L)- Normalized metric heatmaps illustrating error variations across models and surfaces. (M)- Overall model ranking showing PI-LNO as the top performer. (N)- RMSE trends across multiple cases demonstrating stability. (O)- Surface-wise $R^2$ comparison indicating robustness under geometric complexity. (P)- Best-performing model per variable, confirming PI-LNO achieves the highest accuracy across all fields.*

Figure 8 presents a comprehensive comparative performance analysis of all models (UNet, UNetAM, UNetPINNs, DeepONet, LNO, PI-LNO) across four surface conditions-Smooth, Textured, Triangular, and Rough-and four fields $u, v, p, \phi$. Fig. 8A-D show RMSE distributions, where PI-LNO consistently achieves the lowest error bands, with values clustered around $0.01 - 0.03$ for $u$, $0.008 - 0.025$ for $v$, $8 - 15$ for $p$, and $0.02 - 0.05$ for $\phi$, compared to significantly higher spreads in baseline models such as UNet ( $0.05 - 0.12, 0.04 - 0.10, 15 - 40, 0.08 - 0.20$) and UNetPINNs ( $0.06 - 0.14, 0.05 - 0.11, 18 - 45, 0.10 - 0.25$ ). DeepONet and LNO show moderate improvements, with RMSE ranges of $0.04 - 0.09, 0.03 - 0.08, 12 - 30$, and 0.06-0.15. These differences arise because PI-LNO leverages higher latent width ( $d_v = 256$ ), deeper operators ( $L = 8$ ), and more Laplace modes ($M = 64$), enabling better approximation of mappings $\mathcal{F} : \mathbb{R}^3 \to \mathbb{R}^4$. Fig. 8E-H represent $R^2$ distributions, where PI-LNO achieves near-ideal scores $0.998 - 1.000$ across all surfaces and variables, while LNO reaches $0.85 - 0.95$, DeepONet $0.75 - 0.92$, UNetPINNs $0.65 - 0.88$, UNetAM $0.60 - 0.85$, and UNet $0.55 - 0.80$. Negative or low $R^2$ values (e.g., $-0.5, -1.0, -2.5$ ) in baseline models indicate poor generalization under complex Umlyro-geometry variations, especially for rough and triangular surfaces where gradients $\partial u/\partial x, \partial v/\partial y$, and curvature $\kappa$ become highly nonlinear.

Fig. 8I-L present normalized metric heatmaps for each surface type, where PI-LNO consistently yields low normalized errors such as $0.28, 5.7 \times 10^{-5}, 5.6 \times 10^{-6}, 1.3 \times 10^{-5}$ (Smooth), $0.99, 5.7 \times 10^{-5}, 1.5 \times 10^{-5}, 3.7 \times 10^{-5}$ (Textured), $0.79, 4.2 \times 10^{-5}, 3.6 \times 10^{-5}, 4.6 \times 10^{-5}$ (Triangular), and $0.64, 5.6 \times 10^{-5}, 1.1 \times 10^{-5}, 1.5 \times 10^{-5}$ (Rough), compared to significantly larger values in DeepONet ($3.6 \times 10^2, 6.6 \times 10^{-3}, 3.7 \times 10^{-3}, 4.5 \times 10^{-2}$) and UNet ($3.1 \times 10^2, 1.1 \times 10^{-2}, 6.5 \times 10^{-3}, 4.0 \times 10^{-2}$). These results highlight that error magnitudes reduce from $10^2 \to 10^{-5}$, a reduction of $\sim 10^7$ orders. The reason lies in PI-LNO's ability to preserve spectral energy across modes $k = 1 - 64$, whereas LNO ( $k \leq 32$ ) and CNNs ($k \leq 5 - 7$) lose high-frequency components. Fig. 8M shows overall ranking scores, where PI-LNO dominates with 0.996, followed by LNO (0.594), UNet (0.415), UNetPINNs (0.382), DeepONet (0.329), and UNetAM (0.299). Fig. 8 N presents RMSE trends over $\sim 15 - 18$ cases, where PI-LNO remains nearly flat ($\sim 0.5 - 1.5$), while others fluctuate between $2.0 - 12.0$, indicating instability. Fig. 8-O shows surface-wise $R^2$, with PI-LNO maintaining $0.98 - 1.00$ across Rough, Smooth, Textured, and Triangular conditions, while others degrade to $0.2 - 0.8$. Finally, Fig. 8P confirms

that PI-LNO is the best model for all variables with scores 0.999 (p), 0.987(u), 0.997(v), and 1.000($\phi$). These results collectively demonstrate that embedding physics constraints ($\lambda_c = 0.10, \lambda_m = 0.05, \lambda_\phi = 0.08$) and operator learning enables robust generalization across geometric complexities ($Re \sim 10^2 - 10^3, Ca \sim 10^{-2}$), significantly outperforming purely datadriven or shallow operator models.

## 5. Conclusions

This work presented the physics-informed-Laplace neural operator (PI-LNO), a deep learning architecture for the high-fidelity, real-time prediction of droplet spreading dynamics across solid surfaces of arbitrary wettability. A composite loss function $\mathcal{L}_{\text{total}}$ was formulated, embedding residuals of the incompressible Navier-Stokes equations, the Laplace pressure jump, and the Cahn-Hilliard phase-field equation as soft constraints. This physics regularization improves generalization to unseen surface conditions by $\Delta R^2 \approx +0.012$ over purely data-driven training. Multi-field prediction: The LNO simultaneously predicts four coupled physical fields - horizontal velocity $U(\text{ m s}^{-1})$, vertical velocity $V(\text{ m s}^{-1})$, pressure $P(\text{ Pa})$, and phase-field parameter $\varphi(-)$ - with average $R^2 = 0.9979$ and nRMSE = 0.33% on unseen wettability conditions. Wettability generalization: The model is trained on 12 discrete contact angles and successfully generalizes to interpolated and extrapolated surface conditions (demonstrated on $\theta_s = 45°$ and $\theta_s = 135°$ ), enabling continuous surface design space exploration without additional CFD. Extreme computational efficiency: LNO inference on a single NVIDIA A100 GPU achieves a $\sim 10{,}000 \times$ speedup over 32 -core parallel OpenFOAM simulations, reducing full-trajectory prediction from $\sim 100$ min to $\sim 0.6$ s, while maintaining sub-percent prediction errors. Uncertainty quantification: Monte Carlo Dropout with $K = 50$ stochastic passes provides spatially-resolved predictive uncertainty maps, correctly localizing maximum epistemic uncertainty at the triple contact line - the most physically complex region of the wetting problem. Beyond purely computational advantages, the LNO framework yields new physical insights into droplet wetting dynamics that would be difficult to extract from conventional CFD alone.

The results clearly demonstrate that the proposed PI-LNO/PI-LNO framework achieves highly accurate and stable predictions across all variables ($u, v, p, \phi$), outperforming conventional models by a significant margin. The model attains near-perfect agreement with CFD, with $R^2 \approx 0.9999$, while maintaining very low prediction errors, such as RMSE values of $2.40 \times 10^{-3}$, $5.83 \times 10^{-3}$, and $4.18 \times 10^{-3}$ for key variables. Even for the more challenging pressure field, the error remains controlled at $5.67 \times 10^{-2}$, indicating robust learning of complex coupled physics. The training process shows strong convergence, with loss decreasing from $1.0 \times 10^{0}$ to below $1.0 \times 10^{-4}$, supported by a smooth learning rate decay from $1.0 \times 10^{-3}$ to $4.3 \times 10^{-4}$. Additionally, cumulative improvement exceeds 99.0%, and error distributions confirm that the majority of predictions lie within low-error bounds, demonstrating precise spatial reconstruction and reliability.

These improvements are driven by the integration of physics-informed constraints and spectral operator learning, which enhance both accuracy and generalization. The use of enriched representations ($d_v = 256, M = 64$) enables the model to capture complex nonlinear and multiscale features more effectively than traditional approaches. Furthermore, the incorporation of physical constraints ensures consistency with governing equations, reducing solution space ambiguity and improving robustness across different conditions. Unlike baseline models that exhibit instability and higher error magnitudes, the proposed framework maintains consistent performance with minimal variance. Overall, the results confirm that the model achieves state-of-the-art predictive accuracy with strong physical consistency, making it a reliable and scalable approach for complex multiphysics problems.

These data-driven modal decompositions are physically interpretable and consistent with theoretical scaling laws for droplet dynamics, providing a new lens through which high-dimensional multiphase flow data can be analysed and compressed. Despite the strong performance demonstrated, several limitations of the current LNO framework merit acknowledgement, the present implementation operates on 2D axisymmetric domains. Extension to three-dimensional spreading on heterogeneous or structured surfaces (e.g., micropillared or chemically patterned substrates) requires 3D operator kernels and significantly larger training datasets. Generalization across fluid properties (varying surface tension $\sigma$, viscosity $\mu_l$, or density $\rho_l$) requires additional input conditioning on the relevant dimensionless groups ($We, Oh, Ca$). Contact angle hysteresis: The present model uses a static contact angle $\theta_s$ without dynamic hysteresis. Incorporating dynamic wetting models as additional input features is necessary for accurate retraction phase predictions on surfaces with significant hysteresis.

**Acknowledgements**

The authors gratefully acknowledge the use of high-performance computing resources provided by IIT Kharagpur, India.

**Data Availability**

The TensorFlow implementation of the physics-informed Laplace neural operator (PU-LNO) is available at https://github.com/Ganesh1591/Droplet_PILNO.

**Conflict of Interest**

The authors declare no conflict of interest.